\documentclass[sigconf]{acmart}
\usepackage{subcaption}
\usepackage{color,soul}
\usepackage{graphicx}
\usepackage{soulpos}
\soulregister{\cite}{1}
\usepackage{makecell} 
\usepackage[most]{tcolorbox}

\usepackage{enumitem}
\newlist{codeitem}{itemize}{3}

\setlist[codeitem]{nosep}

\setlist[codeitem,1]{label=•, left=1em,  font=\ttfamily}
\setlist[codeitem,2]{label=•, left=2em,  font=\ttfamily}
\setlist[codeitem,3]{label=–, left=3em,  font=\ttfamily}

\newtcolorbox[auto counter, number within=section]{mybox}[2][]{%
  colback=gray!10, colframe=black, arc=3mm, boxrule=0.5pt,
  title=Box~\thetcbcounter: #2,#1}

\newtcolorbox[auto counter]{mybox2}[2][]{%
  colback=gray!10, colframe=black, arc=3mm, boxrule=0.5pt,
  title=Box~\thesection: #2,#1}

\tcbset{
  timellmstyle/.style={
    enhanced,
    colback=white,
    colframe=black,
    boxrule=0.6pt,
    arc=2mm,           
    left=6pt,right=6pt,top=6pt,bottom=6pt,
    before skip=6pt, after skip=6pt
  }
}

\AtBeginDocument{%
  }

\setcopyright{acmlicensed}
\copyrightyear{2025}
\acmYear{2025}
\acmDOI{XXXXXXX.XXXXXXX}
\acmConference[XAIFIN'25]{5th Workshop on Explainable AI in Finance}{November 15--18, 2025}{Singapore }

\begin{document}

\title{Towards Explainable and Reliable AI in Finance}

\author{Albi Isufaj}
\authornote{Both authors contributed equally to this research.}

\orcid{0009-0005-1606-2665}

\affiliation{%
  \institution{National Institute of Informatics \\
  Graduate University for Advanced Studies}
  \city{Tokyo}
  \country{Japan}
}

\author{Pablo Moll\'{a}}
\authornotemark[1]
\orcid{0009-0009-7776-7617}
\affiliation{%
  \institution{National Institute of Informatics \\
  Graduate University for Advanced Studies}
  \city{Tokyo}
  \country{Japan}
}

\author{Helmut Prendinger}
\orcid{0000-0003-4654-9835}
\affiliation{%
 \institution{National Institute of Informatics  \\
  Graduate University for Advanced Studies}
 \city{Tokyo}
 \country{Japan}}

\renewcommand{\abstractname}{\textsc{ABSTRACT}}
\begin{abstract}

Financial forecasting increasingly uses large neural network models, but their opacity raises challenges for trust and regulatory compliance. We present several approaches to explainable and reliable AI in finance. \emph{First}, we describe how Time-LLM, a time series foundation model, uses a  prompt to avoid a wrong directional forecast. \emph{Second}, we show that combining foundation models for time series forecasting with a reliability estimator can filter our unreliable predictions. \emph{Third}, we argue for symbolic reasoning encoding domain rules for transparent justification. These approaches shift emphasize executing only forecasts that are both reliable and explainable. Experiments on equity and cryptocurrency data show that the architecture reduces false positives and supports selective execution. By integrating predictive performance with reliability estimation and rule-based reasoning, our framework advances transparent and auditable financial AI systems.
\end{abstract}




\maketitle

\section{Introduction}


Artificial intelligence has transformed financial forecasting, but its reliance on opaque neural network architectures raises concerns for trust, accountability, and regulatory compliance. 
While accuracy remains a central performance metric \cite{isufaj2025CCM, ESN, feuerriegel2016news, mai2024transformer}, financial stakeholders require more: they need to know both \emph{when} and \emph{why} a prediction or trade should be acted upon. A wrong forecast in trading can translate directly into monetary loss, and in high-stakes environments, the cost of false positives is particularly severe. 

Legislation such as the EU General Data Protection Regulation (GDPR, Article 22, 2018, Recital
71) and the EU AI Act asserts the right for explainability of decisions, and even to challenge them. This calls for a suitable explanation mechanism that is understandable to human users. Rules encoding the data
(market conditions, price history, technical signals etc.) are a natural candidate, as they can be intuitively
understood. Non-profitable trades are unavoidable in trading; yet, the system needs to be able to reveal its
decision rules and how they were applied.

Explainable AI (XAI) has provided an interesting approach to mitigate these concerns. \cite{xai-in-fin} develops different feature selection strategies in an applied financial setting to predict the next-day returns for stocks.
\cite{ohana2021explainable} utilizes SHAP values to predict strong crash-risk, and simple date-by-date explanations of which features drove the model. \cite{xai-review} offers a systematic review on XAI in finance.



In this paper, we propose some approaches to explainability in financial time series forecasting.

\begin{itemize}
\item Time-LLM \cite{time-llm}, where \emph{Prompt-as-Prefix} interacts with the model’s time series forecaster, by concatenating prompt vectors with the series embedding. The prompt is human-readable, so it’s easy to audit why the model leaned one way.

\item Meta-labeling \cite{deprado}, a kind of ``Corrective AI'', where a primary model makes a forecast and a secondary model acts as a \emph{reliability estimator}.
Alternatively, \emph{symbolic reasoning} can encode domain rules for accept/abstain decisions, providing transparent, rule-based justifications that can be inspected by practitioners and regulators.
\end{itemize}

The remainder of the paper is organized as follows. Section \ref{sec:xai-fin} reviews explainable LLMs for financial time series. Section 3 outlines Time-LLM with Prompt-as-Prefix. Section 4 presents the corrective AI (meta-labeling) with a reliability estimator. Section 5 sketches a knowledge-based rule system. Finally Section 6 concludes with limitations and future work.

\section{Explainable Financial Time Series Forecasting with LLMs}
\label{sec:xai-fin}

\cite{temporaldatameetsllm} propose to use LLMs for financial time series forecasting. Experiments include (1) trying zero-shot/few-shot inference with GPT-4 and instruction-based fine-tuning with a public LLM model Open LLaMA, and (2) a combination of zero-shot/few-shot inference techniques with the state-of-the-art AI model GPT-4, and instruction-based fine-tuning using Open LLaMA.
The primary focus is on forecasting weekly/monthly stock returns (defined as the percentage change in stock price from the beginning to the end of the week/month) with accompanying explanations. This focus aligns well with the expertise of Large Language Models (LLMs).

Using ChatGPT 5 \textit{Thinking}, we asked for a 7-day BTC outlook based only on a daily chart covering the prior 90 sessions (Apr 9--Jul 8, 2025). Box~\ref{box:btc-prompt}.1 shows the natural-language prompt used with ChatGPT 5 \textit{Thinking} version, for 7-day BTC directional forecasting from a single TradingView chart, explicitly constraining the model to chart-only evidence.

{\small
\begin{mybox}{Prompt}\label{box:btc-prompt}
Forget about any conversation we had about BTC price prediction;
As a senior analyst with advanced expertise in technical analysis, predict the Bitcoin (BTC) price trend for the next 7 days by solely examining the provided chart from TradingView. The chart shows 90 days of candlesticks from Apr 9--Jul 8, 2025.

\medskip
\textit{[Instructions]:} Do not consider any historical data, market sentiment, or other external factors beyond what is displayed in the chart.

\medskip
\textit{[Examine the Chart]:}  Look at the starting and ending points of the chart, the price movements, and any visible trends or patterns.

\medskip
\textit{[Predict the Trend]:} Using only the visible data from the chart, predict the BTC price trend and expected price over the next 7 days.

\medskip
\textit{[Reasoning Behind the Prediction]:} Why do you predict the BTC price will move in this way over the next 7 days? Discuss the key factors and technical signals from the chart that support your forecast.
\end{mybox}
}


{\small
\begin{mybox}{ChatGPT 5 \textit{Thinking} RESPONSE:}\label{box:chatgpt-output}

\textbf{7-day call} \\
\textbf{Bias:} mild bullish / range-bound continuation \\
\textbf{Expected 7th-day range:} \$108.5k – \$114.5k \\
\textbf{Most likely 7th-day close:} around \$112k–\$113k (near or a touch above the current ~\$112.1k)

\medskip
\textbf{Why this is my base case (from what’s visible):}
\begin{itemize}
    \item \textbf{Uptrend intact:}  
    Higher lows since the mid-chart dip; no lower-low or topping setup.
    
    \item \textbf{\$110k pivot + compression:}  
    Tight, small-body candles = volatility contraction, favoring grind-up continuation over reversal.
    
    \item \textbf{Resistance overhead is shallow, not dominant.}  
    stalls at \$112–113k without bearish engulfing, so incremental pushes higher remain likely.
    
    \item \textbf{Layered supports below.}  
    Demand wick ~\$100–101k and higher lows \$106–108k cut the odds of a deep 7-day drop unless those levels break.

\end{itemize}

\medskip
\textbf{Scenario map (next 7 sessions)}
\begin{itemize}
    \item \textbf{Base (60\%):} Sideways-to-up drift within \$108.5k–\$114.5k; closes \$112k–\$113k.

    \item \textbf{Upside break (25\%):} Daily close > \$113k invites a push toward \$114.5k–\$115.5k.

    \item \textbf{Pullback (15\%):} Fail at \$112–113k → check \$108k–\$109k; bigger trend still intact unless \$106k gives way.
 
\medskip
\textit{Not financial advice—just a technical read of the chart shown.}
\end{itemize}
\end{mybox}
}

Box~\ref{box:chatgpt-output}.2 is the response from ChatGPT 5 \textit{Thinking} citing visible chart structures (e.g., trends, ranges, support/resistance). In reality, BTC was about \$108k on Jul 8 and closed near \$117k on Jul 15.  ChatGPT 5 \textit{Thinking} predicts correctly the upside trend direction of the prices, with \$114.5k–\$115.5k band prediction and with only ~1.3\% inaccuracy.




\section{Prompt-as-Prefix with Time-LLM} 

\begin{figure*}[tbh!] 
\centering
\begin{subfigure}[t]{0.48\linewidth}
  \centering
  \includegraphics[width= \linewidth]{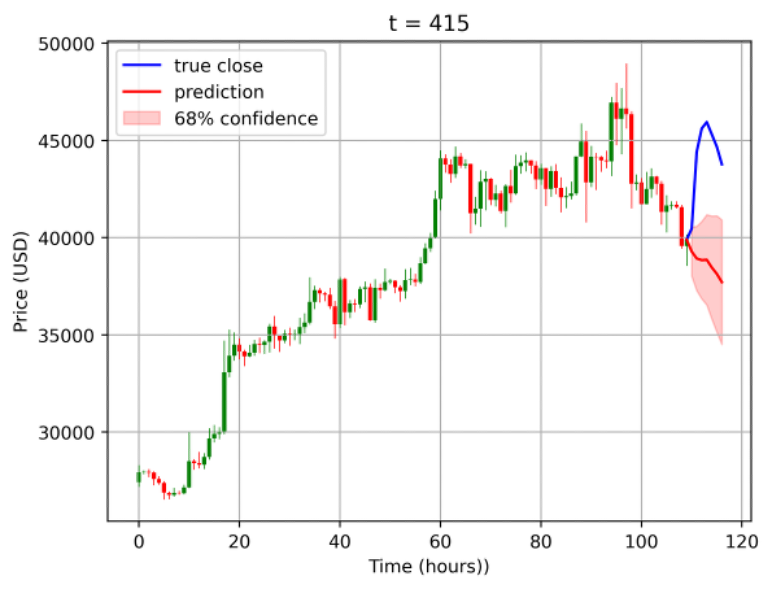}
  \caption{}
\end{subfigure}
\hfill
\begin{subfigure}[t]{0.48\linewidth}
  \centering
  \includegraphics[width=\linewidth]{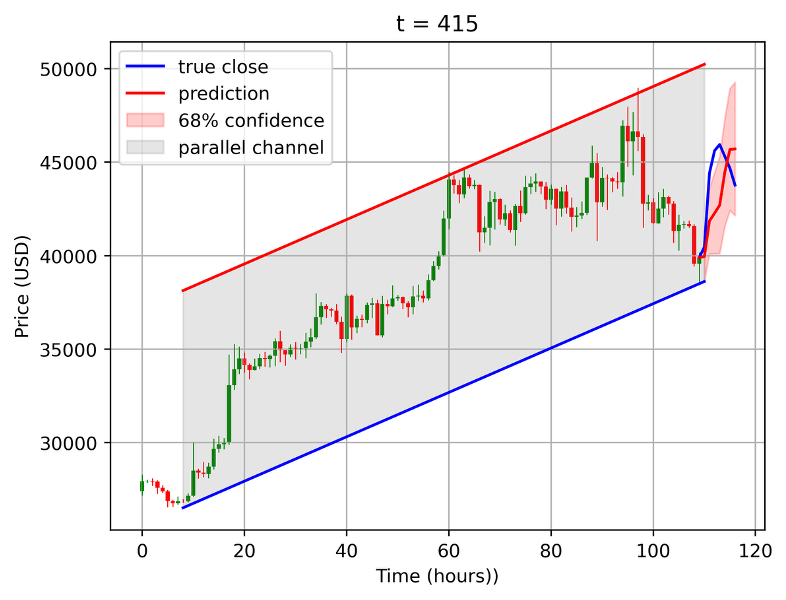}
  \caption{}
\end{subfigure}
\caption{ Time-LLM forecasts for BTC price prediction. (a) Baseline using only OHLC + volume input. (b) With Prompt-as-Prefix  in Box~\ref{box:btc-timeLLM-prompt} and automated created support and resistance line. It controls the forecast toward channel-consistent continuation and tightens the 68\% confidence interval. }
\label{fig:Time-LLM-sr-lines}
\end{figure*}





In \cite{Owen+others.2025}, we build on the Prompt-as-Prefix method introduced in \cite{time-llm} to efficiently add financial context to our input time series. Specifically, we add basic technical analysis, such as automated trend lines and other
technical indicators, to our model.
In this way, we seek to activate the model’s prior knowledge of these indicators
that were likely encountered in its training corpus. Our architecture embeds time series sequences into high-dimensional vectors, concatenates them with prompt
vectors, and inputs the result into a frozen large language model (LLM) \cite{Owen+others.2025}.

Box~\ref{box:btc-timeLLM-prompt} lists a structured prefix used with Time-LLM: domain notes, statistics, and geometric cues (support/resistance) appended ahead of the encoded TS to condition multi-step forecasts. 

Figure \ref{fig:Time-LLM-sr-lines} contrasts Time-LLM without Prompt-as-Prefix or generated support and resistance lines (a) and with it (b), with the prompt yielding a clear improvement due to the prompt that alerts on the support and resistance lines.



{\small
\begin{mybox2}{Prompt}\label{box:btc-timeLLM-prompt}
This dataset is the Bitcoin daily price chart. 
Below is the information about the input time series:

\medskip
\textit{[Domain]:} The bitcoin price is a highly volatile price chart which is globally on an upward trend although it oscillates between bull and bear market cycles that last around 1 to 2 years. Each data point indicates the OHLC price of Bitcoin as well as the volume.
[Instructions]: Predict the data for the next 7 steps given the previous 110 steps.

\medskip
\textit{[Statistics]:} The input has a minimum value of 26511.2 and a maximum value of 49011.4, with an average value of 39621.6.
Your predictions should take into account the behaviour that Bitcoin prices tend to revert when approaching these support and resistance levels.

\medskip
\textbf{1. Support Line}: This sequence represents the lower boundary of the Bitcoin price range over the considered period. Here is the support line : [26511.03 28884.81 31258.58 33632.36 36006.13 38379.9]. It is by definition a line.

\medskip
\textbf{2. Resistance Line}: This sequence represents the upper boundary of the Bitcoin price range over the considered period. Here is the support line : [38130.86 40504.64 42878.41 45252.18 47625.96 49999.73]. It is by definition a line.
\end{mybox2}
}

\section{Reliability Estimation Model as Corrective AI} 

Corrective AI \cite{hands-on-ai-trading} in financial forecasting is an approach where a one model is making a prediction and another model estimates the trustworthiness of the primary model. 

This kind of decision decoupling has been introduced by Lopez de Prado's meta-labeling concept \cite{deprado}. Here, a primary model $M1$ predicts the direction of the price action (up or down), whereas a secondary model $M2$ estimates the reliability of the primary model's prediction.\footnote{Precisely speaking, Lopez de Prado speaks about the \emph{side\/} or a bet versus the \emph{size\/} of a bet. In our work, we want to emphasize the forecasting aspect rather than the trading aspect of the model.} Such approach is important in high-stake decision-making environments (healthcare, automated driving, trading)  with low  tolerance for false positives (FPs), i.e., classifying with a wrong label \cite{selectivenet-2017,Rabanser+others.2025}. In the context of trading, a FP means that our model suggests that, for instance, the price of the asset will go up, while in reality it is not.

\begin{table*}
   \caption{Comparison Table for SPY }
   \label{tab:results-spy}
   \begin{tabular}{lcccccc}   
     \toprule
     Models&Side&Accuracy&Precision&Recall&\makecell[c]{F1\\score}&Execution Rate\\
     \midrule
     Chronos & Up&46\% &55\%&52\%&53\% & 100\%\\
     Chronos + CTTS (ours) & Up&43\%&70\%&7\%&13\%& 6\% \\
     \midrule
     Chronos & Down& 45\% &32\% & 46\% & 38\% & 100\%\\
     Chronos + CTTS (ours) & Down & 65\% & 83\% & 4\% & 8\% & 2\% \\
   \bottomrule
 \end{tabular}
\end{table*} 

We evaluated our algorithms using the following classification metrics: Accuracy, Precision, Recall, F1-score. We also calculate Execution Rate, the proportion of predictions endorsed by the final $M1+M2$ decision in the meta-labeling architecture.

\begin{figure*} [tb]
\centering \begin{subfigure}{0.48\textwidth} \centering 
\includegraphics[width=\textwidth]{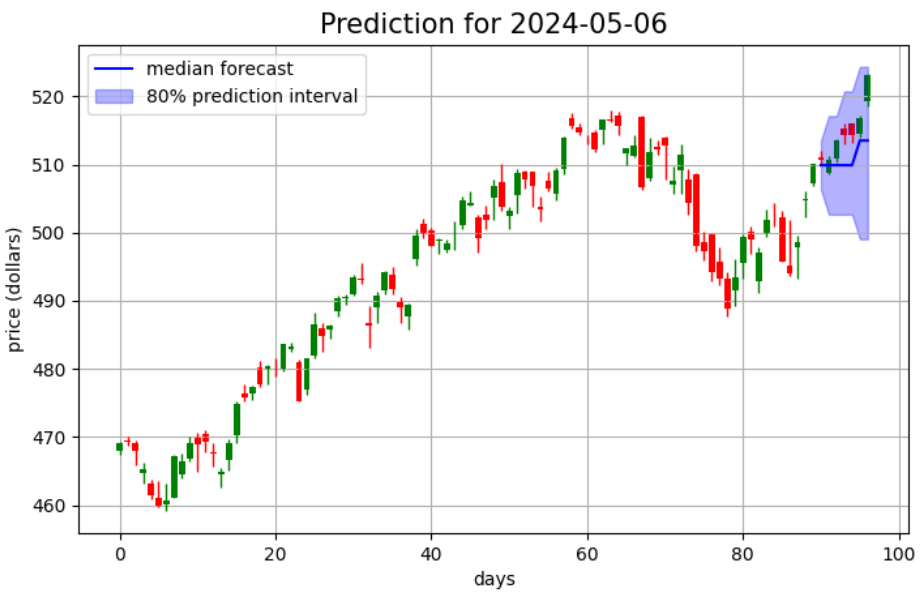} \caption{}
\label{fig:spy-up} \end{subfigure} \hfill{} \vspace{0.1cm}
\centering \begin{subfigure}{0.48\textwidth} \centering 
\includegraphics[width=\textwidth]{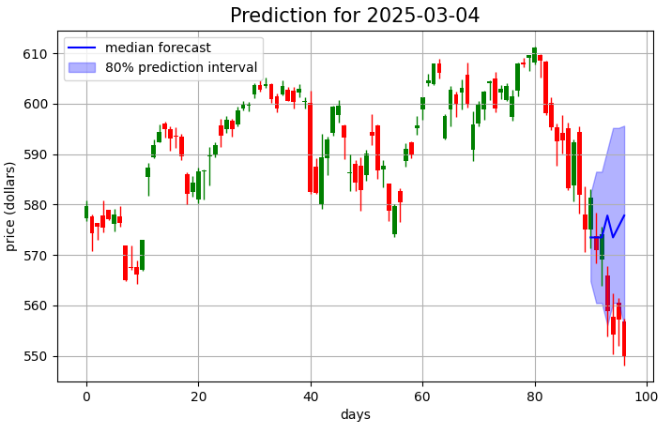} \caption{ }
\label{fig:spy-dn} \end{subfigure} 

\caption{ In Fig.~\ref{fig:spy-up}, $M1$ is predicting `up' for the 7th day and $M2$ confirms $M1$'s reliability (not shown visually), whereas in Fig.~\ref{fig:spy-dn}, $M1$ predicts `up' and $M2$ does not confirm reliability. } 
\Description{Chronos predictions}
\label{fig:chronos-btc} 
\end{figure*}

Table \ref{tab:results-spy}  confirms our hypothesis consistently for SPY for both up trend and down trends. The $M1+M2$ model, i.e., fine-tuned Chronos \cite{chronos}  with Top 20 equities data from S\&P 500 index, form 2001-01-02 to 2024-03-01 with a CTTS \cite{CTTS} enhanced by RevIn \cite{revIn} and Learnable Positional Encoding \cite{attention} reliability model  trained on the predictions of Chronos,  is more precise than $M1$. As expected, the increase in Precision comes at the cost of reducing Recall. Since $M2$ filters out unreliable predictions, the Execution Rate is reduced. This aligns well to the experience in trading where professional traders only execute trades with a high probability of a gain. 

Figure \ref{fig:chronos-btc} shows two examples of regression results of Chronos for SPY closing prices, which constitute two success cases for the meta-labeling architecture.
\begin{itemize}
    \item Fig.~\ref{fig:spy-up} shows a case where $M1$ makes a correct prediction and $M2$ confirms its reliability. 
    \item Fig.~\ref{fig:spy-dn} shows a case where $M1$ makes a wrong prediction and $M2$ prevents execution of prediction as $M1$'s prediction is not reliable.
\end{itemize}

\section{Knowledge-based Reasoning} 



While foundation models and neural predictors provide strong forecasts, they remain opaque and may fail under shifting market regimes. To address this, one could complement them with a symbolic, rule-based reasoning layer \cite{lars,eiter2023logic,le2021scalable} that governs whether forecasts should be accepted, rejected, or deferred. Unlike purely neural corrective models, this layer is grounded in formally defined trading rules and domain knowledge (e.g., candlestick setups, support/resistance conditions). These rules produce human-legible justifications for execution decisions, increasing both operational trust and regulatory compliance, in line with recent calls for explainability in financial AI\cite{xai-review}.

The reasoning layer thus goes beyond filtering unreliable forecasts: it provides an explanation mechanism that can reveal why a prediction was endorsed or rejected, and under what market conditions. Explanations can be framed in intuitive financial terms (``the trade was vetoed because volatility exceeded the safe threshold'') rather than in opaque statistical metrics. 

The integration of knowledge-based reasoning will transform the meta-labeling framework \cite{deprado} into a neurosymbolic system \cite{eiter2023logic}, where execution requires both statistical confidence and logical justification. In doing so, the approach enhances transparency, trust, and robustness in financial AI applications.

As a simple illustrative example, we show the reasoning with a bottoming-tail candlestick detector. The predicate
\texttt{bottoming\_tail\_candle}
holds only if subordinate conditions about relative low, candle size, volume, and tail/body geometry are met over a recent window (e.g., 90 days). Each sub-predicate is a measurable criterion; violating any invalidates the rule. 
\begin{tcolorbox}[colback=white,colframe=black!70,boxrule=0.5pt]
\ttfamily 

bottoming\_tail\_candle \textbf{IF}
    \begin{itemize}
      \item BT\_candle\_is\_lowest\_in\_recent\_history
      \item BT\_candle\_is\_large
      \item BT\_candle\_fulfills\_BT\_criteria
    \end{itemize}
    
BT\_candle\_is\_lowest\_in\_recent\_history \textbf{IF}
    \begin{itemize}
      \item lowest\_candle\_in\_last\_90\_days \% bottom (L) is lowest
    \end{itemize}
    
BT\_candle\_is\_large \textbf{IF}
    \begin{itemize}
      \item candle\_size\_in\_top\_70\%\_of\_last\_90\_candles \% top (H) to bottom (L)
    \end{itemize}
    
BT\_candle\_fulfills\_BT\_criteria \textbf{IF}
    \begin{itemize}
      \item candle\_volume\_in\_top\_10\%\_of\_last\_90\ candles
      \item tail\_at\_least\_50\%\_of\_entire\_candle \% top (H) to bottom (L)
      \item body\_in\_upper\_50\%\_of\_entire\_candle
      \item closing\_price\_in\_top\_25\%\_of\_entire\_candle
    \end{itemize}
\end{tcolorbox}

In practice, these rules can be encoded in a stream-reasoning formalism (e.g., LARS \cite{lars}) and evaluated online, producing a traceable accept/reject rationale. 





 

\section{Conclusions}
\label{sec:conclusions}
The approaches discussed in this paper are aimed to contribute to develop robust and verifiable socio-technical systems, and to move from black-box systems to ones that are transparent, understandable, and reliable. They support transparent decision-making in high-stake applications, including
finance, energy systems, and climate modeling. In that, explainability is a key issue.
Beyond technical gains, the presented approaches contribute to governance and regulatory compliance by offering transparent, auditable decision rationales. 

Future work will explore richer rule-based reasoning, adaptive evaluation under regime shifts, and integration with real-world trading workflows. 


\begin{thebibliography}{21}


\ifx \showCODEN    \undefined \def \showCODEN     #1{\unskip}     \fi
\ifx \showISBNx    \undefined \def \showISBNx     #1{\unskip}     \fi
\ifx \showISBNxiii \undefined \def \showISBNxiii  #1{\unskip}     \fi
\ifx \showISSN     \undefined \def \showISSN      #1{\unskip}     \fi
\ifx \showLCCN     \undefined \def \showLCCN      #1{\unskip}     \fi
\ifx \shownote     \undefined \def \shownote      #1{#1}          \fi
\ifx \showarticletitle \undefined \def \showarticletitle #1{#1}   \fi
\ifx \showURL      \undefined \def \showURL       {\relax}        \fi
\providecommand\bibfield[2]{#2}
\providecommand\bibinfo[2]{#2}
\providecommand\natexlab[1]{#1}
\providecommand\showeprint[2][]{arXiv:#2}

\bibitem[Ansari et~al\mbox{.}(2024)]%
        {chronos}
\bibfield{author}{\bibinfo{person}{Abdul~Fatir Ansari}, \bibinfo{person}{Lorenzo Stella}, \bibinfo{person}{Caner Turkmen}, \bibinfo{person}{Xiyuan Zhang}, \bibinfo{person}{Pedro Mercado}, \bibinfo{person}{Huibin Shen}, \bibinfo{person}{Oleksandr Shchur}, \bibinfo{person}{Syama~Syndar Rangapuram}, \bibinfo{person}{Sebastian~Pineda Arango}, \bibinfo{person}{Shubham Kapoor}, \bibinfo{person}{Jasper Zschiegner}, \bibinfo{person}{Danielle~C. Maddix}, \bibinfo{person}{Michael~W. Mahoney}, \bibinfo{person}{Kari Torkkola}, \bibinfo{person}{Andrew~Gordon Wilson}, \bibinfo{person}{Michael Bohlke-Schneider}, {and} \bibinfo{person}{Yuyang Wang}.} \bibinfo{year}{2024}\natexlab{}.
\newblock \bibinfo{title}{Chronos: Learning the Language of Time Series}.
\newblock
\urldef\tempurl%
\url{https://arxiv.org/abs/2403.07815}
\showURL{%
\tempurl}


\bibitem[Beck et~al\mbox{.}(2018)]%
        {lars}
\bibfield{author}{\bibinfo{person}{Harald Beck}, \bibinfo{person}{Minh Dao-Tran}, {and} \bibinfo{person}{Thomas Eiter}.} \bibinfo{year}{2018}\natexlab{}.
\newblock \showarticletitle{LARS: A logic-based framework for analytic reasoning over streams}.
\newblock \bibinfo{journal}{\emph{Artificial Intelligence}}  \bibinfo{volume}{261} (\bibinfo{year}{2018}), \bibinfo{pages}{16--70}.
\newblock


\bibitem[Carta et~al\mbox{.}(2021)]%
        {xai-in-fin}
\bibfield{author}{\bibinfo{person}{Salvatore Carta}, \bibinfo{person}{Alessandro~Sebastian Podda}, \bibinfo{person}{Diego Reforgiato~Recupero}, {and} \bibinfo{person}{Maria~Madalina Stanciu}.} \bibinfo{year}{2021}\natexlab{}.
\newblock \showarticletitle{Explainable AI for financial forecasting}. In \bibinfo{booktitle}{\emph{International Conference on Machine Learning, Optimization, and Data Science}}. Springer, \bibinfo{pages}{51--69}.
\newblock


\bibitem[Chaffard et~al\mbox{.}(2025)]%
        {Owen+others.2025}
\bibfield{author}{\bibinfo{person}{Owen Chaffard}, \bibinfo{person}{Pablo Moll{\'a}}, \bibinfo{person}{Marc Cavazza}, {and} \bibinfo{person}{Helmut Prendinger}.} \bibinfo{year}{2025}\natexlab{}.
\newblock \showarticletitle{Enhancing Large Language Models for Bitcoin Time Series Forecasting}.
\newblock \bibinfo{journal}{\emph{Knowledge-Based Systems}}  \bibinfo{volume}{330} (\bibinfo{year}{2025}), \bibinfo{pages}{114449}.
\newblock


\bibitem[De~Prado(2018)]%
        {deprado}
\bibfield{author}{\bibinfo{person}{Marcos~Lopez De~Prado}.} \bibinfo{year}{2018}\natexlab{}.
\newblock \bibinfo{booktitle}{\emph{Advances in financial machine learning}}.
\newblock \bibinfo{publisher}{John Wiley \& Sons}.
\newblock


\bibitem[Eiter et~al\mbox{.}(2023)]%
        {eiter2023logic}
\bibfield{author}{\bibinfo{person}{Thomas Eiter}, \bibinfo{person}{Tobias Geibinger}, \bibinfo{person}{Nelson Higuera}, {and} \bibinfo{person}{Johannes Oetsch}.} \bibinfo{year}{2023}\natexlab{}.
\newblock \showarticletitle{A Logic-based Approach to Contrastive Explainability for Neurosymbolic Visual Question Answering.}. In \bibinfo{booktitle}{\emph{IJCAI}}. \bibinfo{pages}{3668--3676}.
\newblock


\bibitem[Feuerriegel and Prendinger(2016)]%
        {feuerriegel2016news}
\bibfield{author}{\bibinfo{person}{Stefan Feuerriegel} {and} \bibinfo{person}{Helmut Prendinger}.} \bibinfo{year}{2016}\natexlab{}.
\newblock \showarticletitle{News-based trading strategies}.
\newblock \bibinfo{journal}{\emph{Decision Support Systems}}  \bibinfo{volume}{90} (\bibinfo{year}{2016}), \bibinfo{pages}{65--74}.
\newblock


\bibitem[Geifman and El-Yaniv(2017)]%
        {selectivenet-2017}
\bibfield{author}{\bibinfo{person}{Yonatan Geifman} {and} \bibinfo{person}{Ran El-Yaniv}.} \bibinfo{year}{2017}\natexlab{}.
\newblock \showarticletitle{Selective classification for {Deep Neural Networks}}. In \bibinfo{booktitle}{\emph{31th Conference on Neural Information Processing Systems (NIPS 20217)}}.
\newblock


\bibitem[Isufaj et~al\mbox{.}(2025)]%
        {isufaj2025CCM}
\bibfield{author}{\bibinfo{person}{Albi Isufaj}, \bibinfo{person}{Caio De~Castro Martins}, \bibinfo{person}{Marc Cavazza}, {and} \bibinfo{person}{Helmut Prendinger}.} \bibinfo{year}{2025}\natexlab{}.
\newblock \showarticletitle{Applying time delay convergent cross mapping to Bitcoin time series}.
\newblock \bibinfo{journal}{\emph{Expert Systems with Applications}}  \bibinfo{volume}{277} (\bibinfo{year}{2025}), \bibinfo{pages}{127125}.
\newblock


\bibitem[Jin et~al\mbox{.}(2024)]%
        {time-llm}
\bibfield{author}{\bibinfo{person}{Ming Jin}, \bibinfo{person}{Shiyu Wang}, \bibinfo{person}{Lintao Ma}, \bibinfo{person}{Zhixuan Chu}, \bibinfo{person}{James~Y Zhang}, \bibinfo{person}{Xiaoming Shi}, \bibinfo{person}{Pin-Yu Chen}, \bibinfo{person}{Yuxuan Liang}, \bibinfo{person}{Yuan-Fang Li}, \bibinfo{person}{Shirui Pan}, {et~al\mbox{.}}} \bibinfo{year}{2024}\natexlab{}.
\newblock \showarticletitle{{Time-LLM}: Time series forecasting by reprogramming {Large Language Models}}. In \bibinfo{booktitle}{\emph{The Twelfth International Conference on Learning Representations (ICLR 2024)}}.
\newblock


\bibitem[Kim et~al\mbox{.}(2021)]%
        {revIn}
\bibfield{author}{\bibinfo{person}{Taesung Kim}, \bibinfo{person}{Jinhee Kim}, \bibinfo{person}{Yunwon Tae}, \bibinfo{person}{Cheonbok Park}, \bibinfo{person}{Jang-Ho Choi}, {and} \bibinfo{person}{Jaegul Choo}.} \bibinfo{year}{2021}\natexlab{}.
\newblock \showarticletitle{Reversible instance normalization for accurate time-series forecasting against distribution shift}. In \bibinfo{booktitle}{\emph{International Conference on Learning Representations}}.
\newblock


\bibitem[Le-Phuoc et~al\mbox{.}(2021)]%
        {le2021scalable}
\bibfield{author}{\bibinfo{person}{Danh Le-Phuoc}, \bibinfo{person}{Thomas Eiter}, {and} \bibinfo{person}{Anh Le-Tuan}.} \bibinfo{year}{2021}\natexlab{}.
\newblock \showarticletitle{A scalable reasoning and learning approach for neural-symbolic stream fusion}. In \bibinfo{booktitle}{\emph{Proceedings of the AAAI Conference on Artificial Intelligence}}, Vol.~\bibinfo{volume}{35}. \bibinfo{pages}{4996--5005}.
\newblock


\bibitem[Mai et~al\mbox{.}(2024)]%
        {mai2024transformer}
\bibfield{author}{\bibinfo{person}{Tai Mai}, \bibinfo{person}{Marc Cavazza}, {and} \bibinfo{person}{Helmut Prendinger}.} \bibinfo{year}{2024}\natexlab{}.
\newblock \showarticletitle{Transformer Models for Bitcoin Price Prediction}. In \bibinfo{booktitle}{\emph{Proceedings of the 2024 3rd International Conference on Frontiers of Artificial Intelligence and Machine Learning}}. \bibinfo{pages}{236--241}.
\newblock


\bibitem[Ohana et~al\mbox{.}(2021)]%
        {ohana2021explainable}
\bibfield{author}{\bibinfo{person}{Jean~Jacques Ohana}, \bibinfo{person}{Steve Ohana}, \bibinfo{person}{Eric Benhamou}, \bibinfo{person}{David Saltiel}, {and} \bibinfo{person}{Beatrice Guez}.} \bibinfo{year}{2021}\natexlab{}.
\newblock \showarticletitle{Explainable AI (XAI) models applied to the multi-agent environment of financial markets}. In \bibinfo{booktitle}{\emph{International Workshop on Explainable, Transparent Autonomous Agents and Multi-Agent Systems}}. Springer, \bibinfo{pages}{189--207}.
\newblock


\bibitem[Pik et~al\mbox{.}(2025)]%
        {hands-on-ai-trading}
\bibfield{author}{\bibinfo{person}{Jiri Pik}, \bibinfo{person}{Ernest~P Chan}, \bibinfo{person}{Jared Broad}, \bibinfo{person}{Philip Sun}, {and} \bibinfo{person}{Vivek Singh}.} \bibinfo{year}{2025}\natexlab{}.
\newblock \bibinfo{booktitle}{\emph{Hands-On AI Trading with Python, QuantConnect, and AWS}}.
\newblock \bibinfo{publisher}{John Wiley \& Sons}.
\newblock


\bibitem[Rabanser et~al\mbox{.}(2025)]%
        {Rabanser+others.2025}
\bibfield{author}{\bibinfo{person}{Stephan Rabanser}, \bibinfo{person}{Anvith Thudi}, \bibinfo{person}{Kimia Hamidieh}, \bibinfo{person}{Adam Dziedzic}, \bibinfo{person}{Israfil Bahceci}, \bibinfo{person}{Akram~Bin Sediq}, \bibinfo{person}{Hamza Sokun}, {and} \bibinfo{person}{Nicolas Papernot}.} \bibinfo{year}{2025}\natexlab{}.
\newblock \showarticletitle{Selective Prediction via Training Dynamics}.
\newblock \bibinfo{journal}{\emph{Transactions on Machine Learning Research}} (\bibinfo{year}{2025}).
\newblock


\bibitem[Sharma et~al\mbox{.}(2025)]%
        {ESN}
\bibfield{author}{\bibinfo{person}{Mansi Sharma}, \bibinfo{person}{Enrico Sartor}, \bibinfo{person}{Marc Cavazza}, {and} \bibinfo{person}{Helmut Prendinger}.} \bibinfo{year}{2025}\natexlab{}.
\newblock \showarticletitle{Echo State Networks for Bitcoin Time Series Prediction}.
\newblock \bibinfo{journal}{\emph{arXiv preprint arXiv:2508.05416}} (\bibinfo{year}{2025}).
\newblock


\bibitem[Vaswani et~al\mbox{.}(2017)]%
        {attention}
\bibfield{author}{\bibinfo{person}{Ashish Vaswani}, \bibinfo{person}{Noam Shazeer}, \bibinfo{person}{Niki Parmar}, \bibinfo{person}{Jakob Uszkoreit}, \bibinfo{person}{Llion Jones}, \bibinfo{person}{Aidan~N Gomez}, \bibinfo{person}{{\L}ukasz Kaiser}, {and} \bibinfo{person}{Illia Polosukhin}.} \bibinfo{year}{2017}\natexlab{}.
\newblock \showarticletitle{Attention is all you need}.
\newblock \bibinfo{journal}{\emph{Advances in neural information processing systems}}  \bibinfo{volume}{30} (\bibinfo{year}{2017}).
\newblock


\bibitem[Yeo et~al\mbox{.}(2025)]%
        {xai-review}
\bibfield{author}{\bibinfo{person}{Wei~Jie Yeo}, \bibinfo{person}{Wihan Van Der~Heever}, \bibinfo{person}{Rui Mao}, \bibinfo{person}{Erik Cambria}, \bibinfo{person}{Ranjan Satapathy}, {and} \bibinfo{person}{Gianmarco Mengaldo}.} \bibinfo{year}{2025}\natexlab{}.
\newblock \showarticletitle{A comprehensive review on financial explainable AI}.
\newblock \bibinfo{journal}{\emph{Artificial Intelligence Review}} \bibinfo{volume}{58}, \bibinfo{number}{6} (\bibinfo{year}{2025}), \bibinfo{pages}{1--49}.
\newblock


\bibitem[Yu et~al\mbox{.}(2023)]%
        {temporaldatameetsllm}
\bibfield{author}{\bibinfo{person}{Xinli Yu}, \bibinfo{person}{Zheng Chen}, \bibinfo{person}{Yuan Ling}, \bibinfo{person}{Shujing Dong}, \bibinfo{person}{Zongyi Liu}, {and} \bibinfo{person}{Yanbin Lu}.} \bibinfo{year}{2023}\natexlab{}.
\newblock \bibinfo{title}{Temporal Data Meets LLM -- Explainable Financial Time Series Forecasting}.
\newblock
\showeprint[arxiv]{2306.11025}~[cs.LG]
\urldef\tempurl%
\url{https://arxiv.org/abs/2306.11025}
\showURL{%
\tempurl}


\bibitem[Zeng et~al\mbox{.}(2023)]%
        {CTTS}
\bibfield{author}{\bibinfo{person}{Zhen Zeng}, \bibinfo{person}{Rachneet Kaur}, \bibinfo{person}{Suchetha Siddagangappa}, \bibinfo{person}{Saba Rahimi}, \bibinfo{person}{Tucker~Hybinette Balch}, {and} \bibinfo{person}{Manuela Veloso}.} \bibinfo{year}{2023}\natexlab{}.
\newblock \bibinfo{title}{Financial Time Series Forecasting using CNN and Transformer}.
\newblock
\urldef\tempurl%
\url{https://arxiv.org/abs/2304.04912}
\showURL{%
\tempurl}


\end{thebibliography}


\end{document}